\documentclass[runningheads]{llncs}
\usepackage[T1]{fontenc}

\usepackage{graphicx,verbatim}

\usepackage{adjustbox}
\usepackage{booktabs}
\usepackage{amsmath}
\usepackage{multirow}

\begin{document}

\title{Leveraging Adversarial Learning for Pathological Fidelity in Virtual Staining}

\author{
José Teixeira\inst{1,2}\orcidID{0009-0004-6917-3003}\and
Pascal Klöckner\inst{2} \orcidID{0009-0005-2381-0908} \and
Diana Montezuma\inst{3}\orcidID{0000-0001-9551-4589} \and
Melis Erdal Cesur\inst{2}\orcidID{0000-0001-8841-1768} \and
João Fraga\inst{3}\orcidID{0000-0001-6053-1709} \and
Hugo M. Horlings\inst{2}\orcidID{0000-0003-4782-8828}\and
Jaime S. Cardoso\inst{1}\orcidID{0000-0002-3760-2473} \and
Sara P. Oliveira\inst{2}\orcidID{0000-0002-6586-9079}
}
% index{Teixeira, José}
% index{Klöckner, Pascal} 
% index{Montezuma, Diana} 
% index{Erdal Cesur, Melis}
% index{Fraga, João}
% index{Horlings, Hugo M.} 
% index{Cardoso, Jaime S.} 
% index{Oliveira, Sara P.} 

\authorrunning{Teixeira \textit{et al.}}

\institute{
INESC TEC, Faculdade de Engenharia, Universidade do Porto, Porto, Portugal \and
The Netherlands Cancer Institute, Amsterdam, The Netherlands\\ \email{s.oliveira@nki.nl} \and
IMP Diagnostics, Porto, Portugal}

\maketitle             

\begin{abstract}
In addition to evaluating tumor morphology using H\&E staining, immunohistochemistry is used to assess the presence of specific proteins within the tissue. However, this is a costly and labor-intensive technique, for which virtual staining, as an image-to-image translation task, offers a promising alternative. Although recent, this is an emerging field of research with 64\% of published studies just in 2024. Most studies use publicly available datasets of H\&E-IHC pairs from consecutive tissue sections. Recognizing the training challenges, many authors develop complex virtual staining models based on conditional Generative Adversarial Networks but ignore the impact of adversarial loss on the quality of virtual staining. Furthermore, overlooking the issues of model evaluation, they claim improved performance based on metrics such as SSIM and PSNR, which are not sufficiently robust to evaluate the quality of virtually stained images. In this paper, we developed CSSP2P GAN, which we demonstrate to achieve heightened pathological fidelity through a blind pathological expert evaluation. Furthermore, while iteratively developing our model, we study the impact of the adversarial loss and demonstrate its crucial role in the quality of virtually stained images. Finally, while comparing our model with reference works in the field, we underscore the limitations of the currently used evaluation metrics and demonstrate the superior performance of CSSP2P GAN.

\end{abstract}

\section{Introduction}
Breast cancer (BC) is a heterogeneous disease with variability in presentation, morphology, molecular characteristics, clinical behavior, and therapeutic response~\cite{TheCancerGenomeAtlasNetwork_2012_nature,goldhirsch_annalsoncology_2013}.
In the pathological workflow, consecutive breast tumor sections are cut and analyzed using hematoxylin and eosin (H\&E), and immunohistochemistry (IHC) stainings. While H\&E allows for the analysis of the general morphology of the tissue, IHC highlights the presence of specific proteins, through a series of antibody and enzymatic reactions~\cite{Magaki_Springer_2019}. Specific immuno markers, such as the Human Epidermal Growth Factor Receptor 2 (HER2), play a key role in the clinical categorization of invasive carcinoma, with numerous studies showing a prognostic and predictive significance~\cite{hacking_2022_Cancers}. However, IHC is a laborious and expensive technique, and virtual staining offers a promising alternative.

Virtual staining is an image-to-image translation task that simulates histological staining by converting from one image domain (e.g., H\&E) to another (e.g., IHC). This process reduces laboratory workload and costs while enabling rapid and accurate diagnosis. As a result, the technique has the potential to decrease diagnosis turnaround time, supporting the goals of the World Health Organization's Global Breast Cancer Initiative~\cite{WHO_initiative_2024}.

\subsubsection{Related work in HER2 virtual staining.} 
Although virtual staining from H\&E to IHC breast biomarkers is a relatively recent field, it is developing rapidly. The research began with the work of Liu \textit{et al.}~\cite{Liu_IEEE_2021_PC-StainGAN} in 2021 and has grown significantly, with 64\% of existing studies being published in 2024~\cite{Klockner_NPJDigitalMedicine_2025}.

In 2022, Liu \textit{et al.}~\cite{Liu_CVPR_2022_BCI} introduced the Breast Cancer Immunohistochemical (BCI) dataset, the first publicly available dataset of registered H\&E and HER2 IHC tile pairs from consecutive breast tissue sections. In addition, the authors developed the Pyramid Pix2Pix model using BCI. Building on the work of Isola \textit{et al.}~\cite{Isola_IEEE_2017_Pix2Pix}, this model adds the Pyramid $\mathcal{L}_{1}$ loss, allowing for improved similarity between generated and real IHC images in the context of consecutive tissue sections, by computing their difference at multiple resolutions, \textit{i.e.}, simulating several octave levels. The intuition behind this loss function can be understood from the observation that two similar but distinct objects appear more alike when viewed from a distance because of the loss of high-frequency details. Therefore, Pyramid $\mathcal{L}_{1}$ loss applies a series of four Gaussian filters, acting as low-pass filters, followed by downsampling, to both real and generated IHC images before computing $\mathcal{L}_{1}$ loss.

In the following year, Li \textit{et al.}~\cite{Li_MICCAI_2023_ASP} released the Multi-IHC Stain Translation (MIST) dataset, the second publicly available dataset of registered H\&E-IHC tile pairs from consecutive breast tissue sections, including ER, PgR, HER2, and Ki67 biomarkers. Furthermore, acknowledging that contrastive learning improves model robustness against label noise~\cite{Ghosh_IEEE_2021,Xue_PMLR_2022}, the authors improved the Pyramid Pix2Pix model by incorporating PatchNCE from Park~\textit{et al.}~\cite{Park_ECCV_2020_CUT} and Adaptive Supervised PatchNCE (ASP) into the generator loss function. Both loss functions aim to maximize mutual information in the embedding space of a network using a patch-based InfoNCE loss~\cite{Oord_ArXiv_2019}. However, while the former penalizes differences between the input H\&E and the generated image, the latter penalizes the difference between the generated image and the target IHC. Moreover, ASP incorporates an adaptive weighting scheme, based on cosine similarity and the training epoch, to downweight the contribution of inconsistent patch locations between the generated and real IHC images in the embedding space, thereby addressing the issue of pixel-level misalignment.

The works of Liu \textit{et al.}~\cite{Liu_CVPR_2022_BCI} and Li \textit{et al.}~\cite{Li_MICCAI_2023_ASP} are a reference in the field, as their models and datasets established the foundation for subsequent research and serve as the standard benchmark for model comparison. However, it is important to note that these datasets contain pairs of H\&E and IHC tiles from consecutive tissue sections. Therefore, when using these datasets, one should not only assume that it is possible to predict cell biomarker expression from H\&E morphology, but also that pairs of images, which do not share the same structure, are highly likely to share the same diagnostic information, even if they present different cells. To address this challenge, most authors develop complex architectures based on conditional Generative Adversarial Networks (cGAN) and sophisticated loss functions, often treating the adversarial loss as an accessory and overlooking its impact on the quality of virtual staining. Moreover, in their pursuit of outperforming previous models, they report improved performance based on the Structural Similarity Index Measure (SSIM) and Peak-Signal-to-Noise-Ratio (PSNR), which we argue are not truly representative of the quality of a virtual staining model.

\subsubsection{Scope.} 
The recent release of the HER2match~\cite{klöckner2025gansvsdiffusionmodels} dataset enables a more accurate and robust assessment of virtual staining models, as it consists of well-registered H\&E-HER2 tile pairs from the same tissue section. Recognizing the limitations of current state-of-the-art (SOTA) methods, we used this dataset to iteratively develop our proposed CSSP2P GAN model, investigating the role of adversarial loss in improving the quality of virtually stained images. As a result, despite being simpler than current SOTA approaches, our model achieves good pathological fidelity. In summary, the main contributions of this study include:
\begin{itemize}
    \item The CSSP2P model, a cGAN framework that takes advantage of adversarial loss to achieve heightened performance in pathological fidelity;
    \item An ablation study to explore the impact of each model component, particularly the adversarial loss, on the quality of the generated images;
    \item A comparison of the classical and perceptual image metrics, highlighting the lack of robustness of SSIM and PSNR when assessing the performance of virtual staining models. 
\end{itemize}

% As Max Planck aptly said, ``Insight must precede application''. Therefore, we have taken a step back, hoping to build enough momentum and address the pressing question: Can cell biomarker expression be predicted from H\&E morphology?

\section{Methodology}
We adopted a cGAN architecture using binary cross-entropy loss (cBCE GAN) as our baseline model. In line with the original GAN formulation, this framework comprises a generator and a discriminator, which are alternately optimized at each training step~\cite{Goodfellow_NIPS_2014}. Unlike previous SOTA approaches, our discriminator receives both the IHC and H\&E images as input. In addition to improving the robustness of the GAN against mode collapse~\cite{Isola_IEEE_2017_Pix2Pix}, this strategy promotes the preservation of semantic information from the H\&E image during transformation to IHC. When performing virtual staining, we assume that biomarker expression at the cellular level can be inferred from the H\&E morphology, suggesting that both modalities share the same pathological information. Consequently, conditioning the discriminator on the input H\&E further addresses the challenge of achieving pathological consistency. To assess the impact of this approach, we trained the same framework with the discriminator receiving only the IHC image (BCE GAN).

Conditioning the discriminator on the H\&E improved performance on LPIPS as well as the classical image similarity metrics SSIM, multiscale-SSIM (MS-SSIM), and PSNR, thus indicating increased similarity with real IHC (Table~\ref{tab:ablation}). Additionally, we observe a significant improvement in Fréchet Inception Distance (FID) and Kernel Inception Distance (KID). Consistent with previous expectations, we confirm that conditioning the discriminator on the input H\&E results in a better representation of IHC semantic information, suggesting shared semantic content between H\&E and IHC. However, conditioning on the H\&E also led to a decrease in Contrast-Structure Similarity (CSS) performance, indicating poorer preservation of H\&E morphology. Furthermore, we observed that cBCE GAN occasionally hallucinated missing regions of connective tissue present in both H\&E and IHC. Although samples were taken from the same tissue section, we found that some H\&E structures were occasionally missing in the IHC pair, which we primarily attribute to mechanical and chemical treatments during re-staining~\cite{Lotz_JMI_2023}. Therefore, we point out that this type of hallucination reflects the nature of the data, making it easier for the generator to fool the discriminator.

\begin{table}[b!]
    \centering
    \renewcommand{\arraystretch}{1.1}
    \caption{Ablation study results in the validation set. Higher values are better for metrics with \(\uparrow\), and vice-versa. The standard deviation reflects the variability across evaluated images. The best and second-best metric values are highlighted in bold and underlined, respectively.}
    \begin{adjustbox}{max width=\textwidth}
    \begin{tabular}{lcccccccc} % 9 columns total
        \toprule
        & \multicolumn{3}{c}{\textbf{Human Perception}} & \multicolumn{3}{c}{\textbf{Image Similarity}} & \textbf{Noise} \\
        \textbf{Model} & LPIPS \(\downarrow\) & FID \(\downarrow\) & KID \(\downarrow\) & SSIM \(\uparrow\) & MS-SSIM \(\uparrow\) & CSS \(\uparrow\) & PSNR \(\uparrow\) \\
        \midrule
        BCE & 0.47 ± 0.07 & 73 & 4.5 ± 0.8 & 0.47 ± 0.12 & 0.43 ± 0.12 & \textbf{0.65 ± 0.07} & 19 ± 4 \\
        cBCE & 0.46 ± 0.07 & \textbf{34} & \textbf{1.2 ± 0.6} & \textbf{0.50 ± 0.12} & 0.46 ± 0.12 & \underline{0.41 ± 0.10} & \textbf{20 ± 4} \\
        P2P & \textbf{0.45 ± 0.07} & 50 & 3.1 ± 0.8 & 0.49 ± 0.12 & \underline{0.47 ± 0.14} & 0.40 ± 0.10 & \textbf{20 ± 4} \\
        CSSP2P & \textbf{0.45 ± 0.07} & \underline{44} & \underline{2.7 ± 0.6} & 0.49 ± 0.12 & \textbf{0.48 ± 0.12} & 0.41 ± 0.10 & \textbf{20 ± 4} \\
        \bottomrule
    \end{tabular}
    \end{adjustbox}
    \label{tab:ablation}
\end{table}

To address the issue of missing tissue regions, we extended the cBCE GAN with Pyramid $\mathcal{L}_{1}$ loss~\cite{Liu_CVPR_2022_BCI}, resulting in P2P GAN. As previously mentioned, this loss function addresses the issue of pixel-level mismatch by penalizing differences between generated and real IHC images at different octave levels, thereby reducing the rate of hallucinations. The improved performance on LPIPS and MS-SSIM (Table~\ref{tab:ablation}) suggests a closer similarity between the generated and real images. However, the use of Pyramid $\mathcal{L}_{1}$ loss led to a decrease in performance in FID and KID. In addition, although we were able to reduce the frequency of hallucinations of missing connective tissue, we observed a decrease in CSS performance, indicating poorer preservation of H\&E structures. 

Therefore, building on the definition of the CSS metric~\cite{Liu_IEEE_2021_PC-StainGAN}, we introduced CSS loss into P2P GAN, resulting in our proposed CSSP2P GAN model. This loss function penalizes structural discrepancies between H\&E and IHC, thereby improving the preservation of H\&E morphology. It is defined as,

\begin{equation*}
    \mathcal{L}_{\text{CSS}}(\mathbf{X}, \hat{\mathbf{Y}}) =
    - \frac{1}{\small{\text{B}}}
    \mathbf{1}_{\small {\text{B}}}^\top\log
    \left[
    \frac{1}{\small{\text{C·H·W}}}
    \mathbf{1}_{\small{\text{CHW}}}^{\top}
    \left(\left(\frac{2 \sigma_{\small{\mathbf{X}\hat{\mathbf{Y}}}} + \epsilon}{\sigma_{\small{\mathbf{X}}}^2 + \sigma_{\small{\hat{\mathbf{Y}}}}^2 + \epsilon} + 1 \right) / \ 2 \right)
    \mathbf{1}_{\small{\text{CHW}}}
    \right]
    \mathbf{1}_{\small{\text{B}}},
\end{equation*}

\noindent
where $\epsilon$ is a stabilization constant, $\sigma_{\mathbf{X}\hat{\mathbf{Y}}}$ is the covariance between $\mathbf{X}$ and $\hat{\mathbf{Y}}$, $\mathbf{1}$ are vectors of ones to compute the mean value, with $B$, $C$, $H$, $W$ corresponding to the batch, channel, height and width dimensions. Although we observed some improvements in CSS, the most significant improvements were in MS-SSIM and human perception metrics (Table~\ref{tab:ablation}). Despite being developed specifically to penalize morphological differences between the H\&E and generated IHC, the loss demonstrated a positive regularizing effect during training. We reason that by encouraging the generator to preserve the structures of the H\&E, the loss regularizes the training, forcing the generator to learn more robust image features.

\subsubsection{Implementation Details}
The discriminator network consists of a 4-layer PatchGAN~\cite{Isola_IEEE_2017_Pix2Pix} with spectral normalization, which has been shown to improve the quality of the generated images and accelerate convergence~\cite{Miyato_ArXiv_2018_Spectral_Normalization}. For the generator, we use a U-Net~\cite{Ronneberger_MICCAI_2015_Unet} architecture with a ResNet-50~\cite{He_ArXiv_ResNet_2015} backbone pre-trained on ImageNet~\cite{Deng_CVPR_2015_ImageNet}. While batch normalization is applied in the discriminator to stabilize training, as supported by previous studies~\cite{Huang_IEEE_2017,Ioffe_ArXiv_2015_BatchNorm}, we employ instance normalization in the generator, following empirical evidence that it outperforms batch normalization in feedforward style transfer tasks~\cite{Huang_IEEE_2017,Ulyanov_ArXiv_2017_InstanceNorm}.

During validation, we monitor the Learned Perceptual Image Patch Similarity (LPIPS) value between the generated and real IHC images to save the best checkpoint. This metric measures similarity based on the distance between intermediate activations of a pre-trained network and has been shown to closely align with human perception of similarity~\cite{Zhang_2018_IEEE_LPIPS}.

We conducted all experiments using $200$ epochs and a batch size of $8$. We used the AdamW~\cite{Loshchilov_ArXiv_2019} optimizer with $\beta_1 = 0.5$ and $\beta_2 = 0.999$ for both networks, with a weight decay of $1 \times 10^{-5}$. The learning rate was set to $5 \times 10^{-5}$, and we used an exponential learning rate scheduler with a decay rate of $0.995$. All random processes were initialized with the same seed to ensure reproducibility, and the models were trained on an NVIDIA A100 GPU. The code is available at: https://github.com/Jose-Miguel-Teixeira/CSSP2P-GAN.

For model comparison, we trained Pyramid Pix2Pix~\cite{pp2p_git} and ASP~\cite{asp_git} under the same conditions using the authors' GitHub code. However, since the authors did not describe any checkpointing strategy, we used the same approach used during model development. All hyperparameters were selected based on the values reported in the respective articles, when available, or the default values provided in the code. The only exception was the number of epochs, which we adjusted to the size of the used, ensuring that the total number of training steps was equal to those of the original work.

\subsubsection{Dataset}
The HER2match~\cite{klöckner2025gansvsdiffusionmodels} dataset consists of $17$ Whole Slide Images of breast tissue stained with H\&E, followed by re-staining with HER2 (IHC) after de-staining, resulting in pairs of samples from the same tissue section. After curation and registration, the authors tiled and split the dataset at the per-patient level, ensuring that each set contained at least one case from each HER2-score. This resulted in $11,610$ tiles for training, $3,582$ for validation, and $5,980$ for testing. All tiles have a size of $1024 \times 1024$ pixels and a resolution of $0.25\mu m/\text{px}$.

\section{Results and Discussion}

\begin{table}[b]
    \renewcommand{\arraystretch}{1.1}
    \centering
    \caption{Comparison results in the test set. Higher values are better for metrics with \(\uparrow\), and vice-versa. The standard deviation reflects the variability across evaluated images. The best metric values are highlighted in bold, and the second-best are underlined.}
    \begin{adjustbox}{width=\textwidth}
    \begin{tabular}{lccccccccc}
        \toprule
         & \multicolumn{3}{c}{\textbf{Human Perception}} & \multicolumn{3}{c}{\textbf{Image Similarity}} & \textbf{Noise} \\
        \textbf{Model} & LPIPS \(\downarrow\) & FID \(\downarrow\) & KID \(\downarrow\) & SSIM \(\uparrow\) & MS-SSIM \(\uparrow\) & CSS \(\uparrow\) & PSNR \(\uparrow\) \\
        \midrule
        PPix2Pix & \textbf{0.46 ± 0.06} & \underline{67} & \underline{6.4 ± 0.9} & \textbf{0.51 ± 0.10} & \textbf{0.54 ± 0.11} & \underline{0.38 ± 0.10} & \underline{20 ± 3} \\
        ASP & 0.47 ± 0.05 & 102 & 8.7 ± 0.8 & \textbf{0.51 ± 0.10} & \underline{0.50 ± 0.11} & \textbf{0.61 ± 0.09} & \textbf{21 ± 3} \\
        CSSP2P & \underline{0.46 ± 0.07} & \textbf{33} &  \textbf{2.5 ± 0.7} & 0.43 ± 0.13 & 0.44 ± 0.13 & 0.35 ± 0.10 & 18 ± 4 \\
        \bottomrule
    \end{tabular}
    \end{adjustbox}
    \label{tab:benchmarking}
\end{table}

\begin{figure}[h]
    \centering
    \includegraphics[width=\textwidth]{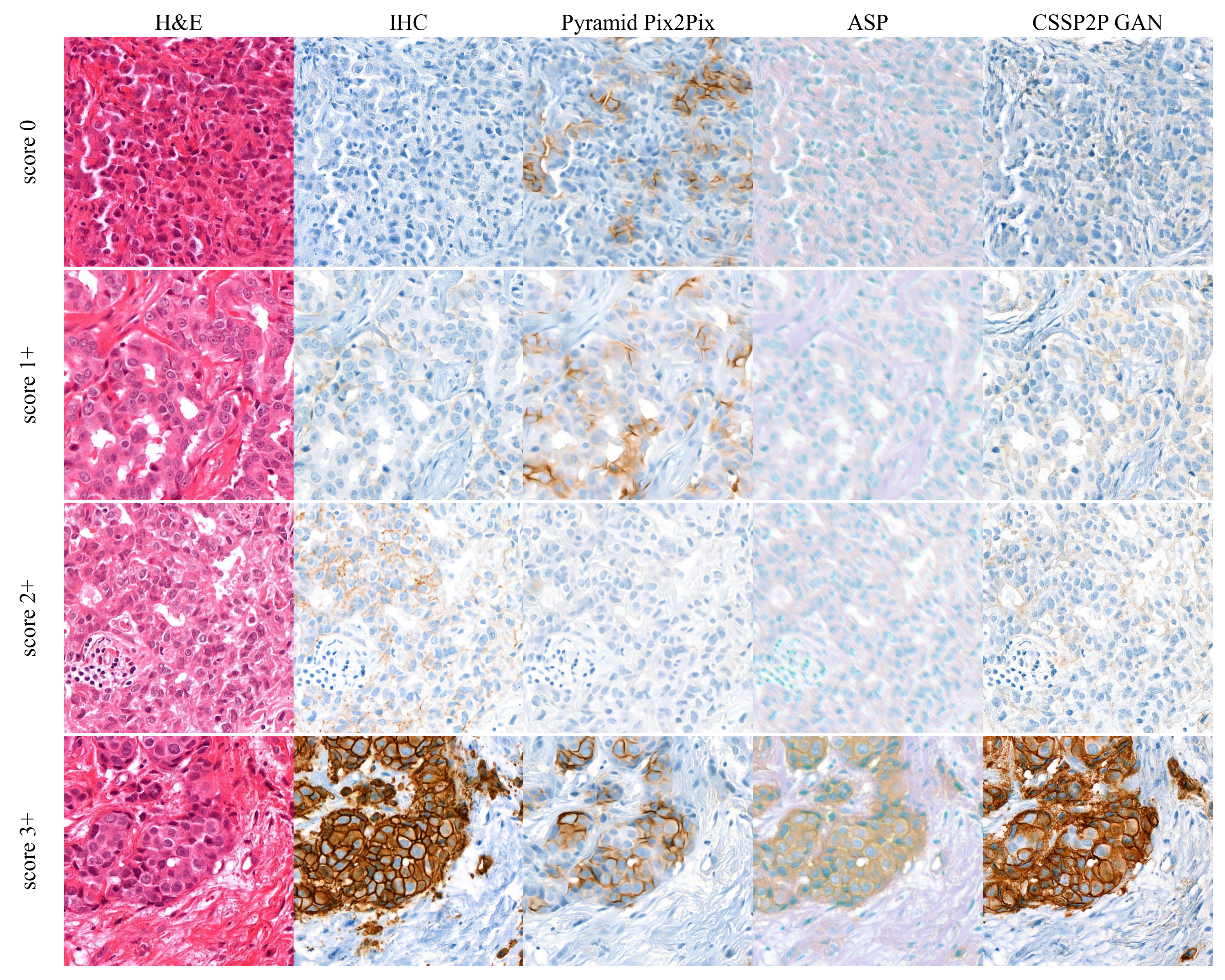}
    \caption{Test set examples, comparing the outputs of Pyramid Pix2Pix, ASP, and CSSP2P GAN models, across the HER2-scores.}
    \label{fig:model_comparison}
\end{figure}

Pyramid Pix2Pix achieved the best LPIPS and MS-SSIM, suggesting a closer similarity to real IHC (Table~\ref{tab:benchmarking}). However, we observe that its outputs are smooth and lack the high-frequency details present in the images generated by CSSP2P GAN (Figure~\ref{fig:model_comparison}). We argue that CSSP2P GAN outputs are perceptually closer to real IHC images. Although both are sampled from the same tissue section, mechanical and chemical manipulations during re-staining can cause pixel-level misalignment between the H\&E-IHC pairs. Consequently, these metrics are less robust in evaluating the similarity between generated and real IHC images, particularly for high-frequency details. Despite appearing in different locations in the generated and real IHC images, these details may still be correct; however, the misalignment can lead to incorrect penalization by these metrics.

Although ASP did not effectively capture the IHC style, its performance was on par with Pyramid Pix2Pix in terms of SSIM and PSNR. This further emphasizes the limitations of the metrics currently used in virtual staining. In fact, ASP performed the worst in FID and KID, suggesting a lower degree of shared semantic information between the real and generated images. Such a result might be due to the competition between the PatchNCE and ASP loss functions. While PatchNCE maximizes the mutual information of a network's embeddings between the generated and input H\&E, ASP uses the same network to maximize the mutual information between the generated image and the real IHC. Therefore, although PatchNCE enabled ASP to excel in CSS, it may have hindered the model from learning the IHC style.

On the other hand, our model significantly outperformed Pyramid Pix2Pix and ASP in both FID and KID. The evidence that the outputs of CSSP2P GAN and real IHC images share the greatest amount of semantic information suggests that CSSP2P GAN is the best model for achieving pathological consistency. However, since these metrics rely on ImageNet pre-trained weights, they cannot be considered definitive indicators of pathological fidelity.

\subsubsection{Qualitative evaluation}
To evaluate the quality and pathological fidelity of the generated images, we invited three pathologists to participate in a blinded assessment. In this experiment, pathologists were presented with real and generated images from CSSP2P GAN side-by-side and asked to answer the following questions: 
\begin{enumerate}
    \item Do the images exhibit a similar staining pattern / diagnostic information?
    \item Which image appears to be of better quality?
    \item Which image is the real one? 
\end{enumerate}

In this experiment, we randomly selected 500 images from the test set, ensuring they were evenly distributed among the HER2 scores. In addition, to minimize observer bias, we randomly included 1\% of extra duplicated images. The layout of the real and generated images was shuffled (left or right), and experts were instructed to use the same monitor size to ensure the same level of detail. To ensure the reliability of the qualitative assessment, the reported values are based on a consensus, \textit{i.e.}, having at least two pathologists agreeing (Table~\ref{tab:pathology_assessment}).

\begin{table}[b!]
    \centering
    \renewcommand{\arraystretch}{1.05}
    \caption{Qualitative evaluation results. The reported values are based on a consensus, which means that at least two pathologists must agree. For questions (2) and (3), the values indicate how often the generated image was selected for each answer.}
    \begin{adjustbox}{width=\textwidth}
        \begin{tabular}{lccccc}
        \toprule
        & \multicolumn{5}{c}{\textbf{HER2-Score}} \\
         & \ \ \ \ \ \textbf{0} \ \ \ & \ \ \ \textbf{1+} \ \ \ & \ \ \ \textbf{2+} \ \ \ & \ \ \ \textbf{3+} \ \ \ & \textbf{All} \\
         \midrule
         (1) Consistent staining pattern? & 0.683 & 0.726 & 0.650 & 0.658 & 0.676 \\
         \hline
         (2) Which image has better quality? & 0.233 & 0.234 & 0.228 & 0.203 & 0.226 \\
         \hline
         (3) Which one is the real image? & 0.683 & 0.661 & 0.585 & 0.602 & 0.636 \\
         \bottomrule
    \end{tabular}
    \end{adjustbox}
    \label{tab:pathology_assessment}
\end{table}

Although the generated images are rated as having lower quality on average, they are often mistaken for real images. This result may suggest that authenticity is linked to specific features rather than overall image quality. In addition to closely resembling authentic IHC images, the result on staining pattern similarity corroborates the expectation that higher performance in preserving semantic information leads to an improved pathological consistency, underscoring the promising performance of our model in achieving pathological fidelity. The quality of CSSP2P GAN is further highlighted by its consistent performance across the HER2-scores.

\section{Conclusion}
In this study, we iteratively developed CSSP2P GAN, and demonstrated the importance of adversarial loss in virtual staining. Specifically, we show that conditioning the discriminator on the input H\&E significantly improves performance on FID and KID. In addition to suggesting better preservation of semantic information between the generated and real IHC images, this result also suggests that this information is shared between H\&E-IHC pairs. Additionally, our pathological evaluation demonstrated that CSSP2P GAN achieved good pathological consistency and closely resembled authentic IHC. Consequently, these results bring us one step closer to answering whether it is possible to predict HER2 expression from H\&E morphology.

Furthermore, this study highlights the shortcomings of the currently used metrics in virtual staining. While MS-SSIM evaluates SSIM across multiple scales, LPIPS assesses differences in the intermediate activations of a network trained on human similarity judgment. As a result, these metrics demonstrate greater robustness to pixel-level misalignment than SSIM and PSNR. However, their results still do not fully align with the perception of quality. 

As future work, we underscore the pressing need to develop a more robust metric to assess the quality of virtually stained images, as well as the availability of more data with H\&E-IHC pairs from the same tissue section, to help with model evaluation and generalizability.

\begin{credits}
\subsubsection{\ackname} This publication is part of the project ``stAIns: AI-based virtual immuno staining from H\&E slides'' with file number NGF.1609.241.018 of the research programme NGF AiNED XS Europa 2024, which is (partly) financed by the Dutch Research Council (NWO). 

\subsubsection{\discintname}
All authors declare no financial or non-financial competing interests.
\end{credits}

% ---- Bibliography ----
\bibliographystyle{splncs04}
\bibliography{Paper-0034.bib}

\end{document}